\documentclass{article}
\usepackage{ijcai26}

\usepackage{times}
\usepackage{soul}
\usepackage{url}
\usepackage[hidelinks]{hyperref}
\usepackage[utf8]{inputenc}
\usepackage[small]{caption}
\usepackage{graphicx}
\usepackage{amsmath}
\usepackage{amsthm}
\usepackage{booktabs}
\usepackage{algorithm}
\usepackage{algorithmic}
\usepackage[switch]{lineno}

\usepackage{amsfonts}
\usepackage{multirow}
\usepackage{array}
\usepackage{subcaption}
\usepackage{float}

\title{Fine-tuning Pre-trained Vision-Language Models in a Human-Annotation-Free Manner}

\author{
Qian-Wei Wang$^{1,2}$\and
Guanghao Meng$^{1,2}$\and
Ren Cai$^{3,2}$\and
Yaguang Song$^2$\And
Shu-Tao Xia$^{1,2}$ \\
\affiliations
$^1$Tsinghua Shenzhen International Graduate School, Tsinghua University\\
$^2$Institute of Perceptual Intelligence, Peng Cheng Laboratory\\
$^3$Peking University Shenzhen Graduate School, Peking University\\
\emails
\{wanggw21, menggh22\}@mails.tsinghua.edu.cn,
2101112278@stu.pku.edu.cn,\\
songyg01@pcl.ac.cn,
xiast@sz.tsinghua.edu.cn
}

\begin{document}

\maketitle

\begin{abstract}
Large-scale vision–language models (VLMs) such as CLIP exhibit strong zero-shot generalization, but adapting them to downstream tasks typically requires costly labeled data. Existing unsupervised self-training methods rely on pseudo-labeling, yet often suffer from unreliable confidence filtering, confirmation bias, and underutilization of low-confidence samples. We propose Collaborative Fine-Tuning (CoFT), an unsupervised adaptation framework that leverages unlabeled data through a dual-model, cross-modal collaboration mechanism. CoFT introduces a dual-prompt learning strategy with positive and negative textual prompts to explicitly model pseudo-label cleanliness in a sample-dependent manner, removing the need for hand-crafted thresholds or noise assumptions. The negative prompt also regularizes lightweight visual adaptation modules, improving robustness under noisy supervision. CoFT employs a two-phase training scheme, transitioning from parameter-efficient fine-tuning on high-confidence samples to full fine-tuning guided by collaboratively filtered pseudo-labels. Building on CoFT, CoFT+ further enhances adaptation via iterative fine-tuning, momentum contrastive learning, and LLM-generated prompts. Extensive experiments demonstrate consistent gains over existing unsupervised methods and even few-shot supervised baselines.
\end{abstract}

\section{Introduction}

In recent years, large-scale pre-trained vision-language models (VLMs) \cite{li2022blip,liu2023visual,bai2025qwen2} such as CLIP \cite{radford2021learning} have revolutionized cross-modal understanding by leveraging massive amounts of general-domain data to learn robust alignments between visual and textual representations. These models have demonstrated remarkable versatility in zero-shot and few-shot settings across diverse downstream tasks, including image classification, retrieval, and visual reasoning. Their ability to generalize from pre-trained knowledge to unseen tasks has made them a cornerstone of modern AI research, enabling applications in fields ranging from medical imaging to autonomous driving.  

However, despite their strong zero-shot performance, fine-tuning remains a critical step to unlock their full potential on specific downstream tasks. Traditional fine-tuning paradigms, whether full-shot (using large labeled datasets) or few-shot (relying on limited annotations), share a common limitation: they depend on manually labeled task-specific data. This reliance on human annotations is not only labor-intensive and costly but also impractical for domains where labeled data is scarce (e.g., rare disease diagnosis) or rapidly evolving (e.g., emerging object categories in robotics).

To address this challenge, researchers have explored unsupervised learning strategies that leverage unlabeled data to enhance model performance. Self-training, a prominent unsupervised approach, involves generating pseudo-labels for unlabeled samples using a pre-trained model and iteratively refining the model with these labels. While promising, existing self-training methods for VLMs \cite{huang2022unsupervised,zhang2023unsupervised,menghini2023enhancing,mirza2023lafter,zhang2024candidate,cao2025latteclip} often suffer from two key issues: (1) These methods lack effective mechanisms for exploiting low-confidence samples. Some approaches utilize only a small number of highest-confidence samples per class for fine-tuning, discarding the remaining unlabeled data; this severely limits the utilization of available data and leaves the model reliant on a narrow subset of potentially biased initial pseudo-labels. Other methods iteratively expand the coverage of pseudo-labels, but fail to incorporate effective safeguards against confirmation bias, making them prone to performance degradation. (2) Their fine-tuning processes usually rely on uni-model adjustments, lacking a collaborative mechanism between multiple models that could leverage cross-modal alignment to generate and validate pseudo-labels more robustly.

In this paper, we propose a novel Collaborative Fine-Tuning (CoFT) framework for effectively adapting vision–language models without any human annotations. The core of CoFT lies in a dual-prompt-based, dual-model cross-modal collaboration mechanism, which enables robust pseudo-label generation and validation without relying on manually designed confidence thresholds or prior assumptions about the label noise ratio. Specifically, we introduce a pair of positive and negative prompts to explicitly model the probability that a pseudo-label is clean. The negative prompt acts as a learnable, data-driven similarity reference that provides a sample-adaptive decision criterion, allowing the model to automatically distinguish clean pseudo-labels from noisy ones. Meanwhile, this negative discrimination signal also regularizes the lightweight tunable modules in the visual branch from a complementary perspective, thereby improving the robustness of visual feature adaptation. In addition, our method adopts a two-phase training paradigm that progressively transitions from parameter-efficient fine-tuning to full fine-tuning. In the early phase, only a small set of high-confidence samples is used to update lightweight tunable parameters in both the textual and visual branches, effectively suppressing confirmation bias. In the later phase, the two models collaboratively perform cross-modal pseudo-label generation and filtering over the entire unlabeled dataset, reliably expanding supervision to the majority of samples and enabling full fine-tuning of the visual encoder. This design establishes an effective bridge between conservative high-confidence learning and large-scale task-specific alignment, allowing CoFT to exploit unlabeled data more thoroughly and robustly. Further building upon the proposed framework, we introduce an enhanced variant, termed CoFT+, which incorporates three key improvements: 1. progressive pseudo-label refinement; 2. contrastive representation enhancement; 3. LLM-genertated prompt templates.

Extensive experiments demonstrate that our methods effectively leverage unlabeled data to boost performance, even outperforming fine-tuning with few manually labeled samples. This work highlights the potential of collaborative self-training in VLMs, offering a cost-effective alternative to labeled data-dependent fine-tuning for task-specific adaptation.

\section{Related Work}  
Large-scale vision–language models (VLMs), such as CLIP \cite{radford2021learning}, learn aligned visual and textual representations from massive image–text pairs \cite{li2022blip,liu2023visual,yang2023dawn,bai2025qwen2}, enabling strong zero-shot generalization. To adapt these models to downstream tasks, fine-tuning strategies range from full fine-tuning, which updates all parameters but risks overfitting on small datasets, to parameter-efficient fine-tuning (PEFT) \cite{han2024parameter}, which updates only a small subset of parameters while freezing the backbone. Representative PEFT methods include prompt learning \cite{zhou2022learning,zhou2022conditional,zhou2022learning,jia2022visual,khattak2023maple}, adapters \cite{houlsby2019parameter,yang2024mma,gao2024clip}, and low-rank adaptation (LoRA) \cite{hu2022lora,zhu2024customize,luo2023lcm}, all of which aim to balance adaptation flexibility and parameter efficiency.

Self-training has been widely explored to leverage unlabeled data via pseudo-labels \cite{xie2020self,tarvainen2017mean,sohn2020fixmatch}. In the context of VLMs, recent works exploit cross-modal alignment to generate pseudo-labels without manual annotation. UPL \cite{huang2022unsupervised} utilizes CLIP’s zero-shot predictions to perform unsupervised prompt learning by selecting top-confidence samples, while DEFT \cite{wei2024vision} treats VLMs as noise detectors to identify clean samples under label noise, demonstrating the potential of self-guided training with pre-trained multimodal knowledge.

% Collaborative learning frameworks further aim to reduce pseudo-label noise by encouraging agreement among multiple models or views \cite{blum1998combining,xu2022cross}. However, collaboration between the visual and textual branches within a single VLM remains underexplored. Our CoFT framework addresses this gap by explicitly modeling cross-modal collaboration to generate and filter high-quality pseudo-labels.

\begin{figure*}[htbp]
    \centering
    \includegraphics[width=0.95\textwidth]{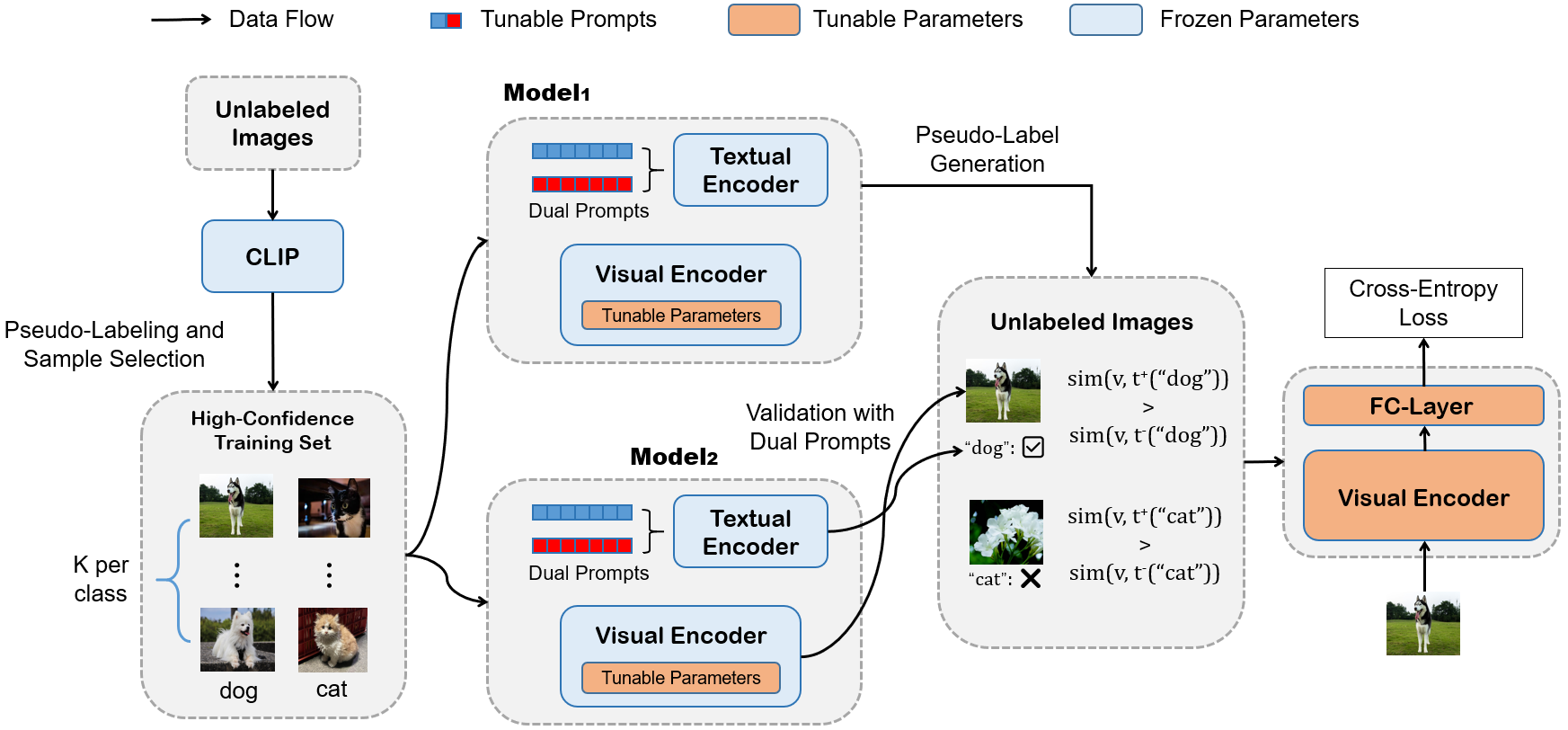}
    \caption{Overall framework of the proposed Collaborative Fine-Tuning (CoFT). CoFT adopts a two-phase collaborative training paradigm to adapt CLIP using only unlabeled data. Phase I: Parameter-Efficient Fine-Tuning (PEFT) with high-confidence pseudo-labels. Pseudo-labels are first generated via CLIP zero-shot inference, from which a small set of high-confidence samples is selected. Two CLIP models are then fine-tuned in parallel using lightweight visual adaptation modules and a dual-prompt learning strategy in the textual encoder, where positive and negative prompts explicitly model the cleanliness of pseudo-labels through a dual-loss objective. Phase II: Collaborative pseudo-label filtering and full fine-tuning. The two models perform bidirectional cross-modal collaboration, alternately generating and validating pseudo-labels over the entire unlabeled dataset using the learned positive–negative prompt similarity criterion. High-quality pseudo-labels are retained to fully fine-tune the visual encoder, enabling robust and scalable task-specific adaptation while mitigating confirmation bias.}
    \label{fig:overall}
\end{figure*}

\section{Methodology}  

Our CoFT framework (see Fig.~\ref{fig:overall}) adopts a two-phase collaborative training paradigm, in which two sub-models are jointly optimized and mutually reinforce each other. In Phase I, we perform parameter-efficient fine-tuning (PEFT) on a small set of high-confidence pseudo-labeled samples to establish reliable task alignment while explicitly preventing early-stage noise amplification (see Section~\ref{phase1}). In Phase II, pseudo-label generation is systematically expanded to the entire unlabeled dataset through bidirectional collaboration between the two task-aligned models. Leveraging their complementary predictions and learned cross-modal discrimination criteria, CoFT performs sample-dependent pseudo-label filtering, retaining only reliable supervision for subsequent training. The resulting high-quality pseudo-labels are then used to fully fine-tune the visual encoder, enabling robust large-scale task adaptation and further boosting task-specific performance (see Section~\ref{phase2}).

\subsection{PEFT with High-Confidence Pseudo-Labels}\label{phase1}

In the first phase, we initialize two CLIP models (denoted as $Model_1$ and $Model_2$) and introduce lightweight trainable parameters into both the textual and visual branches of the models, while keeping the pre-trained parameters frozen. The models are then fine-tuned using a set of highest-confidence initial pseudo-labels obtained from zero-shot inference.

\subsubsection{Pseudo-label generation and high-confidence selection}

We start by generating initial pseudo-labels for the unlabeled downstream images using the pre-trained CLIP model via zero-shot inference. For each image $\boldsymbol{x}$ in the given unlabeled set $\mathcal{U}$, we compute the similarity between its visual embedding $v(\boldsymbol{x})$ encoded by CLIP's visual encoder and textual embeddings of natural language descriptions of all task categories $\{t(\boldsymbol{c}_k)\}_{k=1}^C$ encoded by CLIP's textual encoder, where $v(\cdot)$ and $t(\cdot)$ denote the visual encoder and textual encoder of CLIP respectively, $\boldsymbol{c}_k$ is the natural language description of the $k$-th task category and $C$ is the total number of classes, yielding a probability distribution over classes:
\begin{equation}
    p(\boldsymbol{c}_k \mid \boldsymbol{x}) = \frac{\exp(\text{sim}(v(\boldsymbol{x}), t(\boldsymbol{c}_k)) / \tau)}{\sum_{k'=1}^C \exp(\text{sim}(v(\boldsymbol{x}), t(\boldsymbol{c}_k')) / \tau)}. 
\end{equation}
Here, $\tau$ is a temperature parameter, and $\text{sim}(\boldsymbol{a}, \boldsymbol{b}) = \frac{\boldsymbol{a} \cdot \boldsymbol{b}}{\|\boldsymbol{a}\| \cdot \|\boldsymbol{b}\|}$ denotes cosine similarity. The predicted pseudo-label for each sample is computed as $y = \arg\max\limits_{k} \; p(\boldsymbol{c}_k \mid \boldsymbol{x})$. Then, for each class, we select the top-$K$ samples with the highest confidence scores. These selected samples form the high-confidence subset $\mathcal{S}^*$ with a size of $C \cdot K$, which serves as the training data for the first phase.

\subsubsection{PEFT of visual and textual encoders}

To mitigate the accumulation of errors caused by confirmation bias, we restrict the early-stage optimization to a small set of high-confidence samples and lightweight trainable parameters.

For the textual encoder, we introduce a \textbf{dual prompt learning} mechanism to explicitly model clean and noisy pseudo-labels more robustly. Existing methods for pseudo-label filtering often rely on prior knowledge of the noise ratio or manually designed confidence thresholds, which limits their practical applicability. To overcome this limitation, we maintain a pair of learnable prompts: a positive prompt and a negative prompt.

Formally, for each class $k \in \{1 \dots C\}$, we define:
\begin{align}
    \text{prompt}^+_k = [V^+_1], [V^+_2], ..., [V^+_M], \text{[CLS]}_k, \\
    \text{prompt}^-_k = [V^-_1], [V^-_2], ..., [V^-_M], \text{[CLS]}_k,
\end{align}
where $V^+_i, V^-_i \in \mathbb{R}^d$ are learnable context tokens, $M$ is the total number of tunable textual context tokens and $\text{[CLS]}_k$ denotes the class token.

The positive prompt is trained to uncover discriminative features by maximizing the similarity between image features and their corresponding text features when paired with the true label. In contrast, the negative prompt is designed to behave oppositely: when paired with the true label, its textual representation should be dissimilar to the image features, whereas when paired with a non-true (complement) label, it should exhibit higher similarity. In this way, the negative prompt functions as a learnable, sample-dependent similarity threshold, enabling automatic identification of clean and noisy pseudo-labels without manual heuristics. Meanwhile, the learning of the negative prompt also supervises the lightweight trainable modules in the visual branch from a “negative discrimination” perspective, effectively acting as a form of regularization that improves the robustness of visual adaptation.

On the visual side, we insert lightweight trainable modules to modulate visual representations. In implementation, we adopt Visual Prompt Tuning (VPT) \cite{jia2022visual}, where a small number of learnable visual prompt tokens are prepended to the input sequence of the visual transformer. These prompts are optimized to adapt CLIP’s visual representations to task-specific characteristics without modifying the pre-trained transformer weights, thereby enabling efficient and stable adaptation.

\subsubsection{Dual loss training for dual prompts}

We jointly optimize $Model_1$ and $Model_2$ using two complementary loss functions that correspond to the distinct roles of the positive and negative prompts. This dual-loss design allows the model to both learn discriminative representations from high-confidence samples and identify noisy pseudo-labels in a principled manner.

\begin{itemize}
    \item Loss for Positive Prompts and Visual Prompts ($\mathcal{L}_1$): This loss aligns visual embeddings with their corresponding positive textual embeddings using the high-confidence subset $\mathcal{S}^*$. For a sample $\boldsymbol{x}$ with pseudo-label $y$, we minimize the cross-entropy between the predicted distribution and the pseudo-label:  
    \begin{equation}
    \small
        \mathcal{L}_1 = -\frac{1}{|\mathcal{S}^*|} \sum_{\boldsymbol{x} \in \mathcal{S}^*} \log
        \frac{\exp(\text{sim}(v_{\theta_v}(\boldsymbol{x}), t_{\theta_t}^+(\boldsymbol{c}_y)) / \tau)}{\sum_{k=1}^C \exp(\text{sim}(v_{\theta_v}(\boldsymbol{x}), t_{\theta_t}^+(\boldsymbol{c}_k)) / \tau)},
    \end{equation}
    where $\theta_v$ and $\theta_t$ denote the tuned parameters of visual and textual encoder, $v_{\theta_v}(\boldsymbol{x})$ denotes the visual embedding modulated by visual prompts, and $t_{\theta_t}^+(\boldsymbol{c}_k)$ represents the textual embedding of the positive prompt for class $k$.
    
    \item Loss for Negative Prompts ($\mathcal{L}_2$): This loss ensures that positive prompts are more discriminative than negative prompts for true labels, while the opposite holds for noisy labels. For each sample $\boldsymbol{x}$ with pseudo-label $y$, we randomly sample a complement label $\hat{y}$ from the remaining classes to simulate noise. We formulate the probability of a pseudo-label $y$ being a true label as:
    \begin{equation}
    \tiny
        p^{clean}_{y} = \frac{\exp(\text{sim}(v_{\theta_v}(\boldsymbol{x}), t_{\theta_t}^+(\boldsymbol{c}_y)) / \tau)}{\exp(\text{sim}(v_{\theta_v}(\boldsymbol{x}), t_{\theta_t}^+(\boldsymbol{c}_y)) / \tau) + \exp(\text{sim}(v_{\theta_v}(\boldsymbol{x}), t_{\theta_t}^-(\boldsymbol{c}_y)) / \tau)}.
    \end{equation}
    
    We encourage high-confidence pseudo-labels to be classified as clean, while suppressing the probability that the complement label $\hat{y}$ is clean. The resulting loss is:
    \begin{equation}
        \mathcal{L}_2 = -\frac{1}{|\mathcal{S}^*|} \sum_{\boldsymbol{x} \in \mathcal{S}^*} \log p^{clean}_{y} + \log (1 - p^{clean}_{\hat{y}}).
    \end{equation}
    This loss enforces that $\text{sim}(v_{\theta_v}(\boldsymbol{x}), t^+(\boldsymbol{c}_y)) > \text{sim}(v_{\theta_v}(\boldsymbol{x}), t^-(\boldsymbol{c}_y))$ for true labels and $\text{sim}(v_{\theta_v}(\boldsymbol{x}), t^+(\boldsymbol{c}_{\hat{y}})) < \text{sim}(v_{\theta_v}(\boldsymbol{x}), t^-(\boldsymbol{c}_{\hat{y}}))$ for noisy labels. 
    
\end{itemize}

This objective enforces that, for clean samples, the similarity between the image features and the positive prompt exceeds that of the negative prompt, while the opposite holds for noisy labels. By jointly optimizing $\mathcal{L}_1$ and $\mathcal{L}_2$, the model learns to adapt safely under noisy supervision while preventing error accumulation in the early training stage. The total loss for this phase is $\mathcal{L}_{\text{Phase1}} = \mathcal{L}_1 + \lambda \mathcal{L}_2$, where $\lambda$ balances the two losses. $Model_1$ and $Model_2$ are trained with the same loss but different initialization to encourage diversity.

\subsection{Collaborative Fine-Tuning}\label{phase2}

In this phase, $Model_1$ and $Model_2$ collaborate to generate and filter reliable pseudo-labels from the entire unlabeled set $\mathcal{U}$, which are then used to fully fine-tune the visual encoder.

\subsubsection{Collaborative pseudo-label generation and validation}

The collaboration between $Model_1$ and $Model_2$ is bidirectional, involving mutual pseudo-label generation and validation to improve pseudo-label quality. Without loss of generality, we describe the process where $Model_1$ performs pseudo-label generation and $Model_2$ conducts validation.

\begin{itemize}
    \item \textbf{Pseudo-label generation by $Model_1$:} 
    $Model_1$ generates pseudo-labels for all samples in $\mathcal{U}$ equipped with its fine-tuned textual and visual prompts. For each $\boldsymbol{x} \in \mathcal{U}$, the pseudo-label $y_1(\boldsymbol{x})$ is assigned as the class with the highest similarity to $v_{\theta^1_v}(\boldsymbol{x})$ under $t_{\theta^1_t}^+$, where $\theta^1_t$, $\theta^1_v$ and $\theta^2_t$, $\theta^2_v$ denotes the textual and visual attached parameters of $Model_1$ and $Model_2$, respectively.
    
    \item \textbf{Validation by $Model_2$:} 
    $Model_2$ validates $y_1(\boldsymbol{x})$ using its dual prompts. A sample $\boldsymbol{x}$ with pseudo-label $y_1(\boldsymbol{x})$ is retained in the clean set $\mathcal{S}^1_{\text{clean}}$ if
    $$
    \text{sim}(v_{\theta^2_v}(\boldsymbol{x}), t_{\theta^2_t}^+(\boldsymbol{c}_{y_1(\boldsymbol{x})})) >
    \text{sim}(v_{\theta^2_v}(\boldsymbol{x}), t_{\theta^2_t}^-(\boldsymbol{c}_{y_1(\boldsymbol{x})})).
    $$
    Otherwise, it is assigned to the noise set $\mathcal{S}^1_{\text{noise}}$ and excluded from fully fine-tuning.
\end{itemize}

This collaborative procedure ensures that only samples exhibiting consistent cross-modal alignment, characterized by high positive similarity and low negative similarity, are selected for subsequent supervised training. Finally, we obtain two pairs of noisy and clean subset: $(\mathcal{S}^1_{\text{clean}}, \mathcal{S}^1_{\text{noise}})$ and $(\mathcal{S}^2_{\text{clean}}, \mathcal{S}^2_{\text{noise}})$.

\subsubsection{Fully fine-tuning}

We append a task-specific fully connected (FC) head to the visual encoder to map visual embeddings to class logits:
$f(\boldsymbol{v}) = W_{\text{fc}} \boldsymbol{v} + \boldsymbol{b}_{\text{fc}}$,
where $W_{\text{fc}} \in \mathbb{R}^{C \times d}$ and $\boldsymbol{b}_{\text{fc}} \in \mathbb{R}^C$.
We then fine-tune the entire visual encoder together with the FC head using $\mathcal{S}^{1(2)}_{\text{clean}}$, optimized by the cross-entropy loss over the filtered pseudo-labels:
  
\begin{equation}
\small
\mathcal{L}_{\text{FFT}} =
-\frac{1}{|\mathcal{S}^{1(2)}_{\text{clean}}|}
\sum_{\boldsymbol{x} \in \mathcal{S}^{1(2)}_{\text{clean}}}
\log\left(
\frac{\exp(f(v_{\theta_v}(\boldsymbol{x}))[y^{1(2)}(\boldsymbol{x})])}
{\sum_{k=1}^C \exp(f(v_{\theta_v}(\boldsymbol{x}))[k])}
\right).
\end{equation}

\begin{table*}[ht]
\small
\centering
\setlength{\tabcolsep}{4pt} % 可微调列间距
\begin{tabular}{@{}p{2.0cm}*{8}{p{1.0cm}<{\centering}}@{}}
\toprule
Methods & CLIP & CoFT+ & CoFT & UPL & IFPL & GRIP & LaFTer & CPL \\
\midrule
Caltech101       & 89.27  & 94.40 & \textbf{95.50} & 94.56 & 93.47 & 92.13 & 93.43 & 82.80 \\
DTD              & 44.73  & \textbf{52.30} & 52.13 & 44.68 & 48.24 & 50.90 & 51.77 & 50.11 \\
EuroSAT          & 48.75  & \textbf{90.20} & 86.02 & 66.20 & 39.31 & 59.68 & 45.15 & 63.59 \\
FGVCAircraft     & 25.11  & \textbf{31.25} & 29.49 & 23.67 & 23.10 & 10.14 & 20.91 & 22.86 \\
Flowers102       & 71.73  & \textbf{79.60} & 78.08 & 73.37 & 68.86 & 68.25 & 65.25 & 65.73 \\
Food101          & 84.83  & 90.45 & \textbf{90.58} & 84.71 & 86.55 & 89.16 & 83.72 & 88.64 \\
OxfordPets       & 86.01  & \textbf{93.16} & 92.40 & 90.43 & 89.42 & 91.69 & 87.03 & 90.08 \\
StanfordCars     & 66.52  & \textbf{79.13} & 77.08 & 66.19 & 50.55 & 60.84 & 60.76 & 58.23 \\
UCF101           & 67.44  & \textbf{80.23} & 78.35 & 71.82 & 68.70 & 70.50 & 69.47 & 68.23 \\
\midrule
Average             & 64.93  & \textbf{76.75} & 75.51 & 68.40 & 63.13 & 65.92 & 64.17 & 65.59 \\
\bottomrule
\end{tabular}
\caption{Performance comparison on various benchmark datasets. Best results are shown in \textbf{bold}. All results are reported in percentage (\%).}
\label{tab:main_results}
\end{table*}

\begin{table*}[t]
\centering
\footnotesize
\setlength{\tabcolsep}{6pt}
\begin{tabular}{
  p{1.0cm}
  *{5}{>{\centering\arraybackslash}p{0.8cm}}
  >{\centering\arraybackslash}p{1.0cm}
  >{\centering\arraybackslash}p{1.4cm}
  >{\centering\arraybackslash}p{1.0cm}
  *{3}{>{\centering\arraybackslash}p{1.0cm}}
}
\toprule
Methods & 
CE & 
ELR & 
SCE & 
GMM & 
RoLT & 
UNICON & 
LongReMix & 
ProMix & 
DEFT & 
CoFT &
CoFT+ \\
\midrule
Perf.(\%) & 72.41 & 72.83 & 72.52 & 76.06 & 75.91 & 77.68 & 73.94 & 75.97 & 79.04 & 79.40 & \textbf{80.89} \\
\bottomrule
\end{tabular}
\caption{Performance comparison on CIFAR-100N dataset, whose labels are derived from crowd-sourcing with a noisy rate $r \approx 0.4$.}
\label{tab:cifar100n_results}
\end{table*}

\subsection{Enhanced Collaborative Fine-Tuning}

% Building upon the proposed CoFT framework, we further introduce an enhanced variant, termed CoFT+, which incorporates three complementary improvements to strengthen pseudo-label quality, model diversity, and representation learning. Specifically, CoFT+ extends the first-phase PEFT into multiple iterative rounds, integrates momentum contrastive learning during the second-phase fully fine-tuning, and leverages large language models to generate informative prompt templates for zero-shot initialization.

\subsubsection{Iterative PEFT with progressive pseudo-label refinement}

In CoFT, the first-phase PEFT is conducted only once using high-confidence pseudo-labels obtained from zero-shot inference. In contrast, CoFT+ extends this phase into $R$ iterative rounds of PEFT, enabling progressive refinement of pseudo-labels and deliberate diversification between the two collaborative models.

Concretely, at the $r$-th iteration ($r = 1, \dots, R$), we use the fine-tuned models from the previous iteration, denoted as $Model_1^{(r-1)}$ and $Model_2^{(r-1)}$, to generate pseudo-labels for the unlabeled set $\mathcal{U}$. For each class $k$, we again select the top-$K$ samples with the highest confidence scores, forming two high-confidence subsets $\mathcal{S}^{*(r)}_1$ and $\mathcal{S}^{*(r)}_2$ for the training of $Model_1^{(r)}$ and $Model_2^{(r)}$, respectively.

Unlike conventional self-training, we \emph{re-initialize} both models at each iteration with the original pre-trained CLIP parameters, while retaining only the newly selected subset for training. The models are then fine-tuned via PEFT using the same objectives $\mathcal{L}_1$ and $\mathcal{L}_2$ defined in Section~\ref{phase1}. This design prevents error accumulation from earlier iterations and ensures that each round focuses on a distinct subset of confident samples.

Importantly, although $Model_1^{(r)}$ and $Model_2^{(r)}$ share identical architectures and optimization objectives, they differ in data selection due to stochastic initialization and sampling variations across iterations. This iterative PEFT strategy not only improves pseudo-label accuracy progressively but also increases model diversity from a data-centric perspective, which is crucial for effective collaboration in the subsequent fine-tuning phase.

\subsubsection{Momentum contrastive learning during FFT}

In the second phase, CoFT+ further enhances representation learning by integrating momentum contrastive learning \cite{he2020momentum} alongside supervised fine-tuning with filtered pseudo-labels. This design allows the model to exploit both clean pseudo-labeled samples and the remaining unlabeled or filtered-out samples in a unified manner.

Following the momentum contrast framework, we maintain a momentum visual encoder $v_{\theta_v^m}$ whose parameters are updated as an exponential moving average of the primary visual encoder $v_{\theta_v}$:
\begin{equation}
\theta_v^m \leftarrow \mu \theta_v^m + (1 - \mu) \theta_v,
\end{equation}
where $\mu \in [0,1)$ is the momentum coefficient.

For each image $\boldsymbol{x}$, we generate two stochastic augmented views $\boldsymbol{x}^{(q)}$ and $\boldsymbol{x}^{(k)}$. The query embedding is obtained using the primary encoder,
$
\boldsymbol{q} = v_{\theta_v}(\boldsymbol{x}^{(q)}),
$
while the key embedding is computed using the momentum encoder,
$
\boldsymbol{k} = v_{\theta_v^m}(\boldsymbol{x}^{(k)}).
$

We maintain a queue $\mathcal{Q}$ that stores key embeddings from previous mini-batches. The contrastive loss for a query $\boldsymbol{q}$ is defined as:
\begin{equation}
\small
\mathcal{L}_{\text{Cont}} =
- \log
\frac{\exp(\text{sim}(\boldsymbol{q}, \boldsymbol{k}) / \tau')}
{\exp(\text{sim}(\boldsymbol{q}, \boldsymbol{k}) / \tau') + \sum_{\boldsymbol{k}' \in \mathcal{Q}} \exp(\text{sim}(\boldsymbol{q}, \boldsymbol{k}') / \tau')},
\end{equation}
where $\tau'$ is the temperature parameter for contrastive learning.

During this phase, we jointly optimize the supervised classification loss $\mathcal{L}_{\text{FFT}}$ on the clean pseudo-labeled set and the contrastive loss $\mathcal{L}_{\text{Cont}}$ on all available samples:
\begin{equation}
\mathcal{L}_{\text{Phase2}} = \mathcal{L}_{\text{FFT}} + \gamma \mathcal{L}_{\text{Cont}},
\end{equation}
where $\gamma$ balances supervised and contrastive objectives. This joint optimization encourages class-discriminative representations while preserving instance-level consistency, leading to more robust task-specific visual features.

\subsubsection{LLM-generated prompt templates for zero-shot initialization}

Additionally, CoFT+ employs large language models (LLMs) to automatically generate initial prompt templates for zero-shot inference. Instead of relying on manually designed fixed templates (e.g., ``a photo of a \textless CLASS\_NAME\textgreater''), we prompt an LLM to produce a diverse set of descriptive, task-relevant templates conditioned on the dataset and class semantics.

These LLM-generated templates are used to compute textual embeddings during the initial pseudo-label generation step, providing a stronger zero-shot initialization. Empirically, we observe that more informative prompts yield higher-quality pseudo-labels in early iterations, which in turn benefits both the iterative PEFT phase and the subsequent collaborative fine-tuning. 
%Implementation details of LLM-based prompt templates generation are provided in the appendix.

\begin{figure*}[htbp]
    \centering
    % 第一行
    \includegraphics[width=0.3\textwidth]{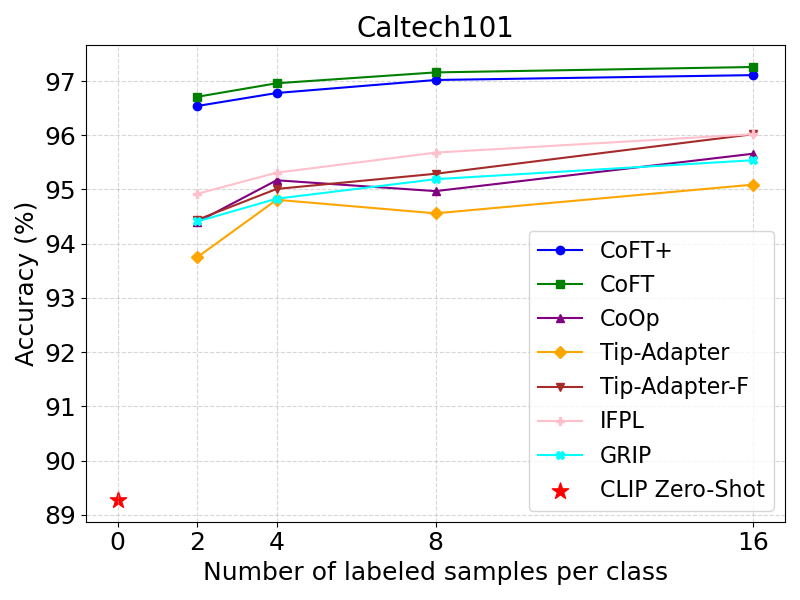}
    \includegraphics[width=0.3\textwidth]{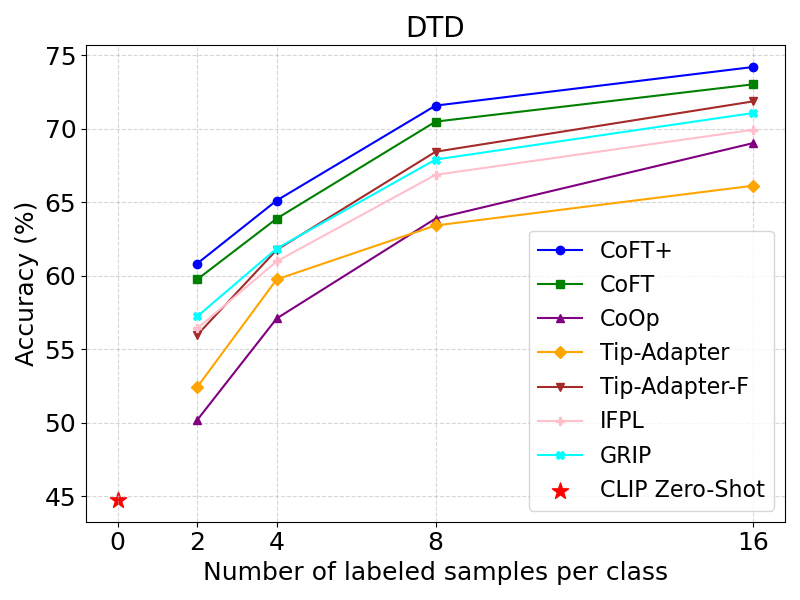}
    \includegraphics[width=0.3\textwidth]{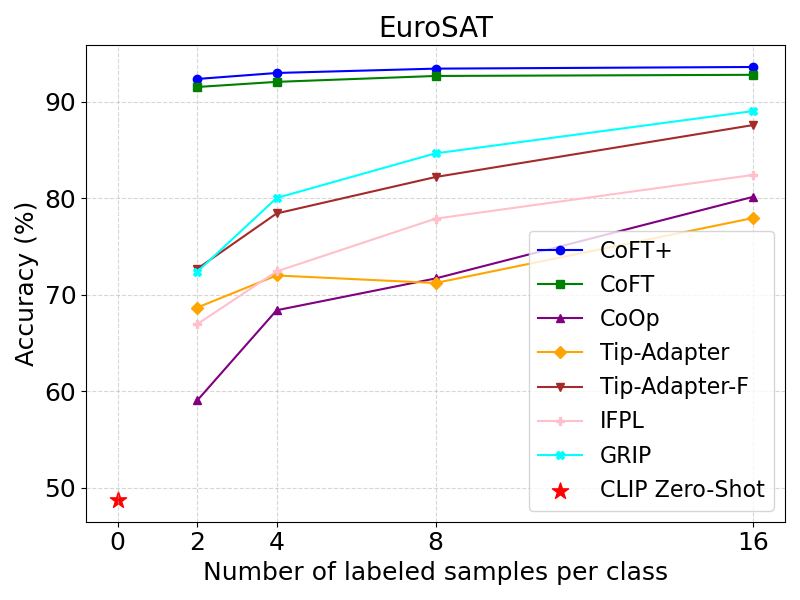}

    % 第二行
    % \includegraphics[width=0.3\textwidth]{./pic/SSL_results/FGVCAircraft_accuracy.png}
    % \includegraphics[width=0.3\textwidth]{./pic/SSL_results/Flowers102_accuracy.png}
    % \includegraphics[width=0.3\textwidth]{./pic/SSL_results/Food101_accuracy.png}

    % 第三行
    \includegraphics[width=0.3\textwidth]{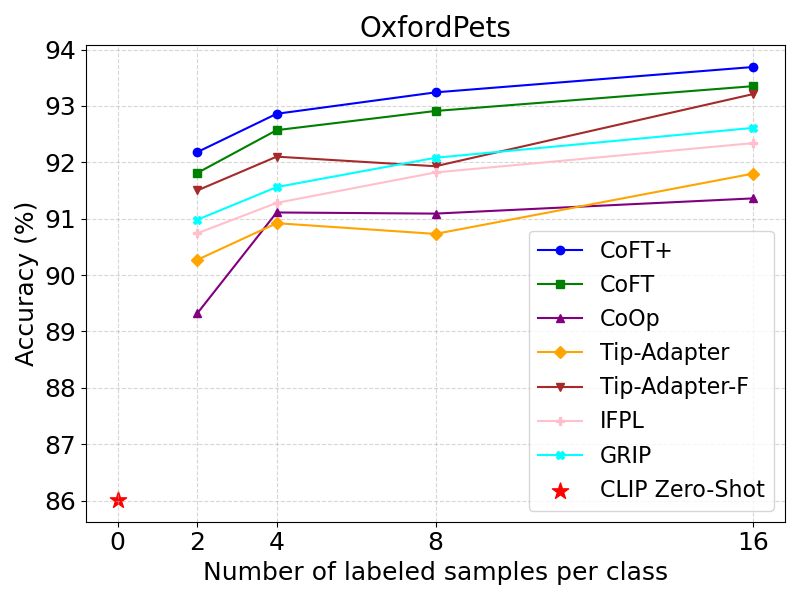}
    \includegraphics[width=0.3\textwidth]{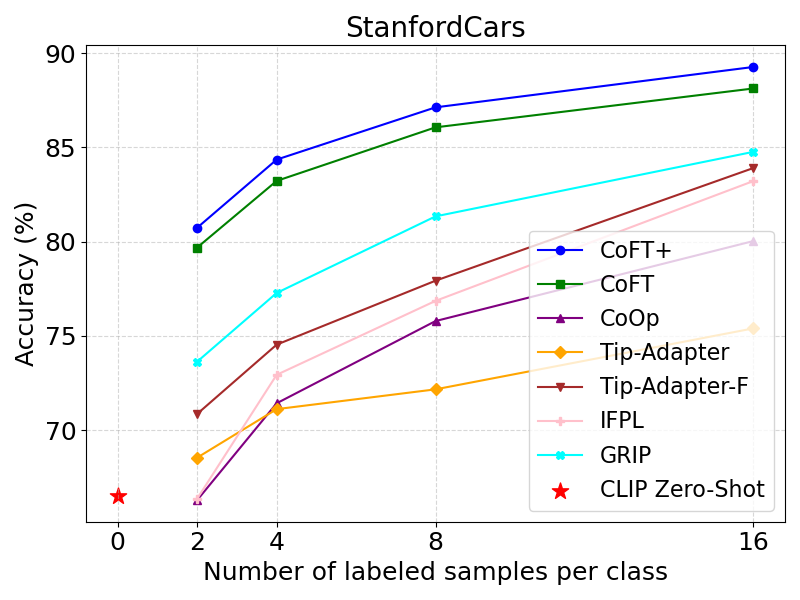}
    \includegraphics[width=0.3\textwidth]{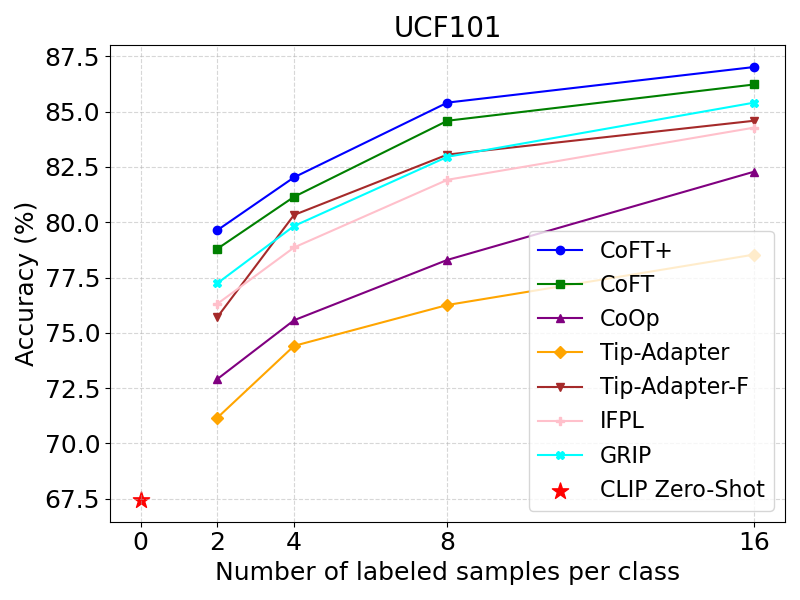}

    \caption{Accuracy comparison of different methods incorporating few-shot labeled samples.}
    \label{fig:SSL_results}
\end{figure*}

\begin{table}[h]
\centering
\small
\setlength{\tabcolsep}{6pt} % 列间距
\begin{tabular}{lcccc}
\toprule
Methods & CLIP & CoFT+ & CoFT & PromptKD \\ \midrule
Settings & ZSL & UL & UL & TRZSL \\
\midrule
Caltech101    & 89.27  & 95.27 & \textbf{95.88}  & 93.61  \\
DTD           & 44.73  & 57.72 & \textbf{57.97}  & 55.08  \\
EuroSAT       & 48.75  & \textbf{89.57} & 87.10  & 63.74  \\
FGVCAircraft  & 25.11  & \textbf{37.67} & 36.86  & 26.24  \\
Flowers102    & 71.73  & \textbf{82.72} & 81.86  & 75.33  \\
Food101       & 84.83  & \textbf{93.53} & 92.59  & 88.84  \\
OxfordPets    & 86.01  & 95.14  & \textbf{95.50} & 91.59  \\
StanfordCars  & 66.52  & \textbf{83.87} & 83.38  & 73.93  \\
UCF101        & 67.44  & \textbf{81.37} & 80.84  & 76.39  \\
\midrule
Average       & 64.93  & \textbf{79.65} & 79.11  & 71.33  \\
\bottomrule
\end{tabular}
\caption{Results of the distillation experiment from CLIP ViT-L/14 to CLIP ViT-B/16. CoFT+ and CoFT use the CLIP ViT-L/14 model in Phase I and fully fine-tune CLIP ViT-B/16 in Phase II. PromptKD uses CLIP ViT-L/14 as teacher and CLIP ViT-B/16 as student in a transductive zero-shot learning (TRZSL) setting, requiring additional ImageNet samples to pre-train the teacher model. ZSL and UL are the abbreviations of zero-shot learning and unsupervised learning, respectively. Best performances are in bold.}
\label{tab:distillation_results}
\end{table}

\section{Experiments}

% \subsection{Experimental setups}
\subsection{Datasets and baselines}

Following previous methods \cite{zhou2022learning,huang2022unsupervised}, we use ten publicly available image classification datasets including Caltech101 \cite{fei2004learning}, CIFAR-10 \cite{krizhevsky2009learning}, DTD \cite{cimpoi2014describing}, EuroSAT \cite{helber2019eurosat}, FGVCAircraft \cite{maji2013fine}, Flowers102 \cite{nilsback2008automated}, Food101 \cite{bossard2014food}, OxfordPets \cite{parkhi2012cats}, StandfordCars \cite{krause20133d}, and UCF101 \cite{soomro2012ucf101}. These datasets cover a variety of different visual classification tasks, from general to fine-grained objects such as texture classification, constituting a comprehensive benchmark.

Except zero-shot inference baseline of CLIP, we compare our methods CoFT and CoFT+ with five unsupervised fine-tuning approaches: UPL \cite{huang2022unsupervised}, IFPL \cite{menghini2023enhancing}, GRIP \cite{menghini2023enhancing}, LaFTer \cite{mirza2023lafter} and CPL \cite{zhang2024candidate} and three few-shot fine-tuning approaches: CoOp \cite{zhou2022learning}, Tip-Adapter \cite{zhang2021tip}, and Tip-Adapter-F \cite{zhang2021tip}. Moreover, in the setting of learning from naturally noisy dataset, our methods are compared with nine learning from noisy labels approaches such as: DEFT \cite{wei2024vision}; and in the setting of distillation, our methods are compared with PromptKD \cite{li2024promptkd}. To ensure a fair comparison, all algorithms were re-evaluated under the same backbone, i.e., ViT-B/16, unless otherwise specified.

% \subsubsection{Implementation details}

% In implementation, we select the top $K=16$ highest-confidence samples for each class for training in the first phase. For training in both phases, we use SGD as the optimizer, with initial learning rates set to 0.03 (for training a small number of tunable parameters) and 5e-4 (for training the entire visual encoder), respectively, along with a cosine decay scheduler. The weight decay is set to 5e-4 and momentum to 0.9. The number of training epochs is selected from \{1, 10, 20\}. For both positive and negative in dual-prompt learning, the number of tunable context tokens is $M=16$, with the class token positioned at the end, and the tunable context tokens for visual branch is $M_v=20$. We conduct all the experiments on one NVIDIA RTX 3090 GPU.

\begin{table*}[htbp]
\centering
\small
\setlength{\tabcolsep}{5pt} % 调整列间距
\begin{tabular}{lcccccccccccc}
\toprule
Dataset & CLIP & CoFT+ & CoFT & \multicolumn{3}{c}{CoOp} & \multicolumn{3}{c}{Tip-Adapter} & \multicolumn{3}{c}{Tip-Adapter-F} \\
\midrule
Settings & ZSL & UL & UL & 2-shot & 4-shot & 8-shot & 2-shot & 4-shot & 8-shot & 2-shot & 4-shot & 8-shot \\
\midrule
Caltech101    & 89.27 & 94.40 & 95.50 & 94.40 & 95.17 & 94.97 & 93.75 & 94.81 & 94.56 & 94.44 & 95.01 & 95.29 \\
% DTD           & 44.73 & 52.30 & 52.13 & 50.18 & 57.09 & 63.89 & 52.42 & 59.75 & 63.42 & 55.97 & 61.76 & 68.44 \\
EuroSAT       & 48.75 & 90.20 & 86.02 & 59.06 & 68.40 & 71.69 & 68.68 & 72.00 & 71.22 & 72.69 & 78.42 & 82.19 \\
% FGVCAircraft  & 25.11 & 31.25 & 29.49 & 27.48 & 30.15 & 37.20 & 31.59 & 32.64 & 36.00 & 33.45 & 36.24 & 42.09 \\
% Flowers102    & 71.73 & 79.60 & 78.08 & 83.96 & 89.32 & 92.16 & 87.29 & 91.76 & 94.03 & 89.69 & 92.85 & 95.45 \\
Food101       & 84.83 & 90.45 & 90.58 & 84.65 & 85.88 & 85.07 & 86.09 & 86.19 & 86.25 & 86.28 & 86.67 & 86.76 \\
OxfordPets    & 86.01 & 93.16 & 92.40 & 89.32 & 91.11 & 91.09 & 90.27 & 90.92 & 90.73 & 91.50 & 92.10 & 91.93 \\
StanfordCars  & 66.52 & 79.13 & 77.08 & 66.30 & 71.43 & 75.79 & 68.55 & 71.12 & 72.17 & 70.87 & 74.54 & 77.93 \\
UCF101        & 67.44 & 80.23 & 78.35 & 72.91 & 75.57 & 78.30 & 71.16 & 74.41 & 76.26 & 75.71 & 80.33 & 83.06 \\
\midrule
Average       & 64.93 & 76.75 & 75.51 & 69.81 & 73.79 & 76.68 & 72.20 & 74.84 & 76.07 & 74.51 & 77.55 & 80.35 \\
\bottomrule
\end{tabular}
\caption{Accuracy comparison of different methods across nine datasets. CoOp, Tip-Adapter, and Tip-Adapter-F are few-shot supervised fine-tuning methods with varying number of shots as indicated, while our methods only require unlabeled samples.}
\label{tab:SSL_results}
\end{table*}

\begin{table*}[htbp]
\centering
\small
\setlength{\tabcolsep}{2pt}
\begin{tabular}{
  >{\centering\arraybackslash}p{1.5cm}| % Backbone
  p{1.8cm}                            % Dataset
  *{8}{>{\centering\arraybackslash}p{1.2cm}} % Methods
}
\toprule
Backbone & Dataset & CLIP & CoFT+ & CoFT & UPL & IFPL & GRIP & LaFTer & CPL \\
\midrule
\multirow{4}{*}{ViT-B/32}
% & Caltech101    & 89.17 & 94.12 & \textbf{94.98} & 93.61 & 92.47 & 91.32 & 93.35 & 82.38 \\
& DTD           & 40.67 & \textbf{51.08} & 50.22 & 43.36 & 46.71 & 48.82 & 46.68 & 47.53 \\
& EuroSAT       & 41.73 & \textbf{81.74} & 80.55 & 64.31 & 38.92 & 57.84 & 73.91 & 62.41 \\
% & FGVCAircraft  & 18.49 & \textbf{24.36} & 23.48 & 22.14 & 21.68 &  9.92 & 17.88 & 21.51 \\
% & Flowers102    & 65.99 & \textbf{75.41} & 74.25 & 72.11 & 67.08 & 66.42 & 71.00 & 65.12 \\
% & Food101       & 77.85 & \textbf{86.04} & 85.96 & 80.14 & 80.21 & 83.32 & 78.72 & 82.58 \\
% & OxfordPets    & 84.95 & \textbf{92.13} & 90.95 & 89.43 & 88.59 & 90.78 & 86.43 & 89.21 \\
& StanfordCars  & 59.49 & \textbf{71.34} & 70.21 & 65.02 & 49.64 & 59.72 & 54.74 & 56.61 \\
& UCF101        & 61.16 & \textbf{75.96} & 74.80 & 70.36 & 67.83 & 69.61 & 68.24 & 67.11 \\
\midrule
\multirow{4}{*}{ViT-L/14}
% & Caltech101    & 92.61 & 96.14 & \textbf{96.61} & 94.91 & 93.71 & 92.94 & 93.43 & 83.12 \\
& DTD           & 55.33 & \textbf{63.11} & 62.30 & 50.82 & 53.61 & 55.21 & 56.03 & 53.79 \\
& EuroSAT       & 59.92 & \textbf{88.93} & 88.18 & 68.13 & 40.31 & 61.48 & 46.79 & 64.21 \\
% & FGVCAircraft  & 36.09 & \textbf{43.12} & 42.22 & 32.84 & 30.63 & 15.23 & 25.48 & 27.11 \\
% & Flowers102    & 78.66 & \textbf{87.84} & 86.93 & 77.21 & 72.42 & 70.52 & 66.81 & 67.32 \\
% & Food101       & 92.90 & \textbf{95.61} & 94.59 & 89.81 & 91.23 & 92.73 & 88.33 & 91.52 \\
% & OxfordPets    & 93.48 & \textbf{97.12} & 96.87 & 92.43 & 91.85 & 95.21 & 88.74 & 91.21 \\
& StanfordCars  & 77.31 & \textbf{88.43} & 87.39 & 72.61 & 54.23 & 63.12 & 62.03 & 60.53 \\
& UCF101        & 76.24 & \textbf{86.31} & 85.33 & 79.11 & 75.61 & 77.23 & 72.62 & 74.52 \\
\bottomrule
\end{tabular}
\caption{Performance comparison under other CLIP backbones (ViT-B/32 and ViT-L/14).}
\label{tab:backbone_comparison}
\end{table*}

\subsection{Main results}

Table~\ref{tab:main_results} summarizes the classification accuracy under the fully unlabeled setting. Both CoFT and its enhanced variant CoFT+ consistently outperform CLIP zero-shot and prior unsupervised CLIP adaptation methods across all nine benchmarks. CoFT+ achieves the best average accuracy of 76.75\%, yielding a substantial improvement of +11.82\% over the CLIP baseline. Notably, large gains are observed on fine-grained and domain-shifted datasets such as EuroSAT, StanfordCars, and UCF101, demonstrating the effectiveness of CoFT in mitigating domain mismatch without labeled data. Compared with recent unsupervised approaches, our methods achieve clear performance advantages on most datasets, with CoFT+ surpassing the strongest competitor by over 26\% on EuroSAT. These results highlight the benefit of the proposed dual-prompt dual-model collaborative pseudo-labeling mechanism, which explicitly models pseudo-label reliability through cross-model validation and effectively reduces confirmation bias while enabling more comprehensive exploitation of unlabeled data.

\subsection{Results on real-world noisy dataset}

Table~\ref{tab:cifar100n_results} reports results on CIFAR-100N, where labels are obtained via crowd-sourcing with a high noise rate ($r!\approx!0.4$). We compare our methods with representative noisy-label learning approaches, including loss-based methods (CE, SCE, ELR, GMM) and recent sample-selection techniques (RoLT, UNICON, LongReMix, ProMix, DEFT), all of which rely on the provided noisy annotations for training. In contrast, CoFT and CoFT+ completely discard human labels and perform fully annotation-free adaptation. Despite this, CoFT achieves an accuracy of 79.40\%, surpassing all competing methods, including the recent state-of-the-art DEFT (79.04\%). The enhanced CoFT+ further improves performance to 80.89\%, establishing a new state of the art on CIFAR-100N. These results demonstrate that collaborative pseudo-labeling with pre-trained vision–language models can generate more reliable supervision than noisy human annotations, offering a more effective and cost-efficient alternative.

\subsection{Results under semi-supervised settings}

To further investigate the effectiveness of our approach in scenarios with limited labeled data, we evaluate its performance under semi-supervised settings, where the model is adapted by jointly leveraging unlabeled data and a small number of manually labeled samples (2-, 4-, 8-, and 16-shot per class). As illustrated in Fig.~\ref{fig:SSL_results}, both CoFT and CoFT+ consistently outperform existing adaptation methods across all nine datasets and shot configurations. 
%Compared with few-shot adaptation methods that rely solely on limited labeled samples, our approach achieves substantially better performance, indicating that learning discriminative representations from only a handful of labeled examples is inherently challenging, while the incorporation of unlabeled data effectively alleviates this limitation. Moreover, CoFT+ yields modest yet consistent improvements over CoFT, suggesting that enhancing pseudo-label quality and representation learning further benefits semi-supervised adaptation.

\subsection{Experiments of distillation scenario}

We further evaluate CoFT and CoFT+ in a distillation setting, transferring knowledge from a larger CLIP ViT-L/14 model to a compact ViT-B/16 model without any human annotations. Compared with PromptKD (see Table~\ref{tab:distillation_results}), which relies on transductive zero-shot distillation and additional auxiliary data, our methods achieve substantially better performance on all benchmarks. The advantage is particularly pronounced on challenging datasets such as EuroSAT, FGVCAircraft, and StanfordCars, indicating stronger generalization under limited supervision. Overall, CoFT+ achieves the best average accuracy, while CoFT delivers comparable results, confirming that both variants effectively transfer cross-modal knowledge from large pre-trained models to smaller architectures. These results highlight the practicality of CoFT for efficient model compression while maintaining strong recognition performance.

\subsection{Comparison with supervised methods}

We further compare our methods with representative few-shot supervised fine-tuning approaches, including CoOp, Tip-Adapter, and Tip-Adapter-F, across nine benchmark datasets, as summarized in Table~\ref{tab:SSL_results}. These baseline methods rely on varying numbers of labeled samples (2-, 4-, and 8-shot per class), whereas our methods (CoFT and CoFT+) require only unlabeled data. Overall, our methods achieve competitive or superior performance on the majority of datasets. 

% While some supervised methods occasionally surpass our approach on specific datasets—for example, CoOp achieves slightly higher accuracy on Flowers102 with 8-shot supervision—the performance gap is generally small. Notably, CoFT and CoFT+ consistently outperform the few-shot baselines on challenging datasets such as EuroSAT, DTD, and UCF101, demonstrating that unsupervised adaptation with pre-trained vision-language models can produce highly discriminative representations even without access to labeled data.

These results highlight the practical value of our unsupervised framework: it achieves comparable or better accuracy than supervised fine-tuning in most cases while entirely avoiding reliance on costly human annotations.

\subsection{Results under different backbones}

Table~\ref{tab:backbone_comparison} reports the performance of our methods and representative baselines under different CLIP backbones, including ViT-B/32 and ViT-L/14. As shown, both CoFT and CoFT+ consistently achieve superior performance across all datasets and backbone configurations, demonstrating strong robustness to the choice of visual encoder.

% Under the ViT-B/32 backbone, CoFT+ attains the best average performance and outperforms all competing methods on most datasets, with particularly notable gains on challenging benchmarks such as EuroSAT, StanfordCars, and UCF101. CoFT exhibits comparable performance, consistently surpassing prior unsupervised and weakly supervised adaptation approaches.

% When switching to the stronger ViT-L/14 backbone, all methods benefit from improved visual representations; however, the relative advantage of CoFT and CoFT+ remains evident. In this setting, CoFT+ continues to deliver the best overall results, while CoFT closely follows, indicating that our framework scales effectively with more powerful CLIP backbones. These observations confirm that the proposed approach is backbone-agnostic and can reliably leverage stronger pre-trained representations without requiring architectural modifications.

\section{Conclusion}
We propose CoFT, a collaborative fine-tuning framework for annotation-free adaptation of vision–language models that mitigates noise accumulation and confirmation bias via a two-phase training strategy. CoFT introduces a dual-prompt, dual-model collaboration mechanism to perform sample-adaptive pseudo-label filtering and extract reliable supervision from unlabeled data. Extensive experiments across unsupervised, noisy-label, and distillation settings demonstrate that CoFT and CoFT+ consistently outperform unsupervised and few-shot supervised baselines.

\bibliographystyle{named}
\bibliography{ijcai26}

\end{document}